\documentclass{IEEEcsmag}

\usepackage[colorlinks,urlcolor=blue,linkcolor=blue,citecolor=blue]{hyperref}

\usepackage{upmath}

\jvol{Volume 0}
\jnum{XX}
\paper{8}
\jmonth{May/June}
\jname{Computer}
\pubyear{2023}

\usepackage{amsfonts}
\usepackage{caption}
\usepackage{multirow}
\usepackage{booktabs}
\usepackage{amsmath}
\usepackage{moresize}
\usepackage{footnote}
\usepackage[normalem]{ulem}
\usepackage{makecell}
\usepackage{ragged2e}
\usepackage[para,online,flushleft]{threeparttable}
\usepackage[symbol]{footmisc}
\let\labelindent\relax
\usepackage{enumitem}
\usepackage{tikz}
\usepackage{textcomp}
\usepackage{hyperref}
\usepackage{lipsum}

\begin{document}

% \sptitle{Department: Head}
% \editor{Editor: Name, xxxx@email}

\title{A Tale of Two Cities: Data and Configuration Variances in Robust Deep Learning}

\author{Guanqin Zhang, Jiankun Sun, Feng Xu, Yulei Sui}
\affil{University of Technology Sydney, Australia}

\author{H.M.N. Dilum Bandara, Shiping Chen}
\affil{CSIRO's Data61, Australia }

\author{Tim Menzies}
\affil{North Carolina State University, USA}

% \markboth{Department Head}{Paper title}

\begin{abstract}
\justifying
Deep neural networks (DNNs), are widely used in many industries such as image recognition, supply chain, medical diagnosis, and autonomous driving. 
However, prior work has shown the high accuracy of a DNN model does not imply high robustness (i.e., consistent performances on new and future datasets) because the input data and external environment (e.g., software and model configurations) for a deployed model are constantly changing. 
Hence, ensuring the robustness of deep learning is not an option but a priority to enhance business and consumer confidence.
Previous studies mostly focus on the data aspect of model variance. 
In this article, we systematically summarize DNN robustness issues and formulate them in a holistic view through two important aspects, i.e., data and software configuration variances in DNNs. 
We also provide a predictive framework to generate representative variances (counterexamples) by considering both data and configurations for robust learning through the lens of search-based optimization.
\end{abstract}

\maketitle

\chapterinitial{Deep Neural Networks (DNNs)} 
are increasingly deployed into critical systems to solve complex applications, including medical diagnosis~\cite{wang2020deep}, electricity supply chain~\cite{ghaderi2017deep}, and drought resilience~\cite{zhang2021dual}. 
However, despite their wide adoption in many complicated tasks,
DNNs are still far from perfect. Existing learning models often yield imprecise or incorrect outputs for real-world applications due to imperfect data and/or configurations~\cite{xiao2021nondeterministic}. For example, multiple identical training procedures may generate different models with accuracy variances in the presence of various factors including data perturbations (e.g., limited, weakly-labeled, and concept-drifting training samples) and variances caused by software implementation and configurations  (e.g., nondeterministic DL layers, and random weight initialization and floating-point imprecision).
Analyzing DNN robustness issues can help practitioners and researchers build more reliable deep learning systems. 
%\textcolor{blue}{with the awareness of} the performance variances among the deployed model. 

Figure~\ref{tab:existing} shows the experiments on a range of applications with performance fluctuations due to data- and configuration-related perturbations. 
In response to such data and configuration variances, the \emph{robustness} property (i.e., minor modifications to the (future) inputs of DNNs must not alter its outputs) has been extensively studied in the deep learning community. 
Robustness is the most noteworthy correctness property of DNNs. Assuring robustness is critically important to prevent AI systems from environmental perturbations. 

Many previous efforts have limited the DNN robustness to single impacts~\cite{ASE2020pham}. Existing studies mostly focus on data factors affecting the robustness of DNNs. Tripuraneni \textit{et al.} \cite{tripuraneni2021overparameterization} study the high-dimensional asymptotics of random regression under covariate shift; Zhang \textit{et al.} \cite{zhang2020familial} investigate the negative effects of weakly-labeled samples for clustering and proposed a new hybrid representation strategy for familial clustering; Tu \textit{et al.} \cite{tu2020better} found that automatic keyword labeling suffers weakly-labeled issue in bug-fixing commits and recommended to label commits through human+artificial expertise; Shu \textit{et al.} \cite{shu2022omni} argue that well-crafted adversarial samples heavily decrease the identification performance of DNN models and proposed a new Omni solution with the multi-model ensemble. 
Some recent studies also explore other factors affecting robustness. For example, Xiao \textit{et al.} \cite{xiao2021nondeterministic} study the impacts of CPU multithreading on DNN systems; 
Pham \textit{et al.} \cite{ASE2020pham} explored the performance of identical models with random seeds and nondeterminism-introducing factors under different training runs. These recent approaches provide some insights to understand the impacting factors on DNN robustness. 
In this proposal, we aim to systematically study a wider range of impacts of the data and software implementation and configurations, which will guide our later tasks in on-demand mitigating and repairing robustness issues of the state-of-the-art (SOTA) DNNs. 

Fewer studies are conducted on a holistic understanding of the associative effects of a wider range of data and configuration factors. In this paper, we first present several influencing factors for both data and configuration variance in a range of real-world downstream domains in Table~\ref{tab:existing}. 
Then, we study the impacts of combinatorial factors regarding DNN robustness on an image classification task. We observe that combinatorial factors impose composition effects on DNN robustness, which are often ignored in the current work.
Following the observation, we formulate the robustness property by considering the perturbations to both data and configurations. 
Finally, we talk about our current research exploring  a novel search-based variance prediction method for accelerating  the exploration of influential factor
on DNNs.  

% (s) and/or their combinations (composition effects) on the robustness of DNN models.

\section{Performance Variances -- Data and Configurations}

DNN models suffer from performance variances due to imperfect data or configurations, which can produce unreliable learning systems for a range of applications.  

\subsection{\textbf{Impacts of Single Factors}}

\begin{figure*}[h]
    \centering
    \includegraphics[width=\textwidth]{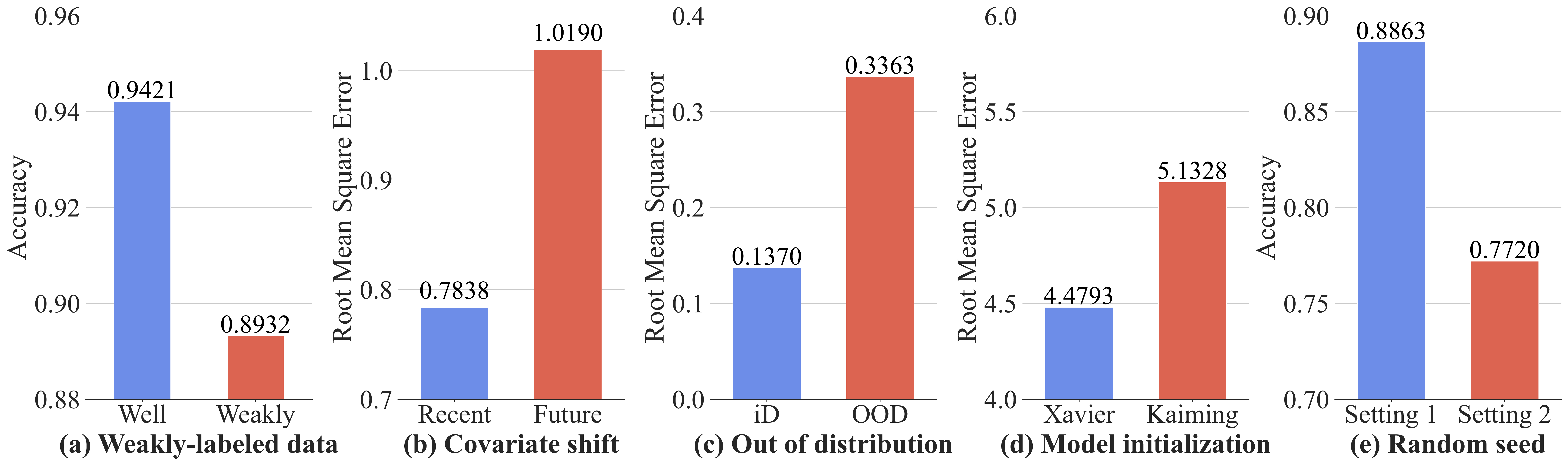}
    \caption{A range of applications in DNN projects with corresponding data and configurations. Accuracy on classification and root mean square error (RMSE) on regression tasks}
    \label{fig:motivation1}
\end{figure*}

Table~\ref{tab:existing} briefly describes five representative DNN projects relating to CSIRO's missions\footnote[2]{CSIRO's missions: Partner with us to tackle Australia’s greatest challenges, \url{https://www.csiro.au/en/about/challenges-missions}}, and Figure~\ref{fig:motivation1} presents the corresponding experimental results on these projects regarding data and configuration perturbations:

\begin{savenotes}
  \begin{table*}[!t]
    \centering
    \caption{A range of applications in DNN projects with their corresponding data and configurations }

 \begin{threeparttable}
    \begin{tabular}{llll}
        \toprule
       \multirow{2}{*}[-0.5\dimexpr \aboverulesep + \belowrulesep + \cmidrulewidth]{\textbf{Application use cases}} &  \multirow{2}{*}[-0.5\dimexpr \aboverulesep + \belowrulesep + \cmidrulewidth]{\textbf{Data}} & \multicolumn{2}{c}{\textbf{Configurations}}    \\
        \cmidrule(lr){3-4} 
         & & \textbf{Network architecture} & \textbf{Learning method} \\
        \midrule

        (a) Recycling waste classification & TrashNet~\tnote{1} & ResNet+Attention~\cite{RecycleNet_trash_images} & SGD \\ 
        (b) Traffic forecasting & Traffic prediction dataset~\tnote{2} & GRU~\cite{TrafficPredictionGRU} & SGD\\
        (c) Water quality forecasting      & Water quality data~\tnote{3}       & Dual HeadSSIM\tnote{6} & Adam   \\
        (d) Renewable energy prediction & Deep-forecast~\tnote{4} & DL-STF~\tnote{7} & RMSprop \\
        (e) Medical diagnosis & Physionet 2017 dataset~\tnote{5} & ResNet~\cite{wang2020deep} & Adam \\ 
        \bottomrule
    \end{tabular}
    \begin{tablenotes}
    \item[1] GitHub, Dataset of trash images, \url{https://github.com/garythung/trashnet} \\
    \item[2] Kaggle, Traffic Prediction Dataset, \url{https://www.kaggle.com/datasets/fedesoriano/traffic-prediction-dataset}\\
    \item[3] Kaggle, Real time water quality data, \url{https://www.kaggle.com/datasets/ivivan/real-time-water-quality-data}\\
    \item[4] Github, Dataset of wind speed, \url{https://github.com/amirstar/Deep-Forecast}\\
    \item[5] The PhysioNet/Computing in Cardiology Challenge 2017, \url{https://physionet.org/content/challenge-2017/1.0.0/}\\
    \item[6] Dual-head sequence-to-sequence imputation model (Dual HeadSSIM ~\cite{zhang2021dual}) denotes the dual-head bidirectional GRU structure with cross-head attention.\\
    \item[7] DL-based Spatio-Temporal Forecasting (DL-STF~\cite{ghaderi2017deep}) denotes the spatio-temporal recurrent neural network. 
  \end{tablenotes}
    \end{threeparttable}
\label{tab:existing}
\end{table*} 
\end{savenotes}

%\begin{itemize}
\begin{enumerate}[label=(\alph*)]
\item {\textbf{Weakly-labeled data}}, which contains incomplete or partially labeled training samples, can adversely affect the accuracy of the trained DNN models. 
The TrashNet dataset contains plastic and glass bottle images with a small proportion of incorrectly labeled data. 
RecycleNet~\cite{RecycleNet_trash_images} utilized in this project has shown a decrease in classification accuracy by 4.89\% due to these weakly-labeled samples in the training set.

\item {\textbf{Covariate shift}}, which demonstrates a distribution shift among different segments of time periods, causes the performance decrease of the trained model. Figure~\ref{fig:motivation1} (b) shows the large variance of RMSE when using two Recurrent Neural Network (RNN) models~\cite{TrafficPredictionGRU} trained with two different temporal data (i.e., 2016.11.1-2017.2.28 and 2017.3.1-2017.6.30) to mitigate the traffic (supply chain) congestion problem. 

\item {\textbf{Out of distribution data}}, which are introduced by the open set, causes the untrustworthy estimation of the DNN model performance. 
Figure~\ref{fig:motivation1} (c) shows the increasing RMSE on the OOD testing set, which reveals the performance degradation of a Gate Recurrent Unit (GRU) model~\cite{zhang2021dual} on the water quality forecasting problem.

\item {\textbf{Model initialization}} causes DNN instabilities varied in different initialization methods. We perform the experiments with the same dataset and implementation, except for the weight initialization in the wind speed forecasting project~\cite{ghaderi2017deep}. 
Figure~\ref{fig:motivation1} (d) depicts the RMSE of the experiment with Kaiming \cite{he2015delving} and Xavier initialization \cite{glorot2010understanding} methods, which yields a difference of around 0.65.

\item {\textbf{Random seed}} introduces non-deterministic factors during the training of DNNs when stochastic algorithms are utilized.
Figure~\ref{fig:motivation1} (e) shows the accuracy differences between two settings for the DNN-based recognition of electrocardiograms (ECGs) in cardiac arrhythmia diagnosis~\cite{wang2020deep}. 
We conducted 30 training jobs with random seeds (where other settings are consistent) on ResNet, and the average performance differences can be over 10\%.
%\end{itemize}
\end{enumerate}

\subsection{\textbf{Impacts of Combinatorial Factors}}

\begin{figure*}[t]
    \centering
    \includegraphics[width=0.73\textwidth]{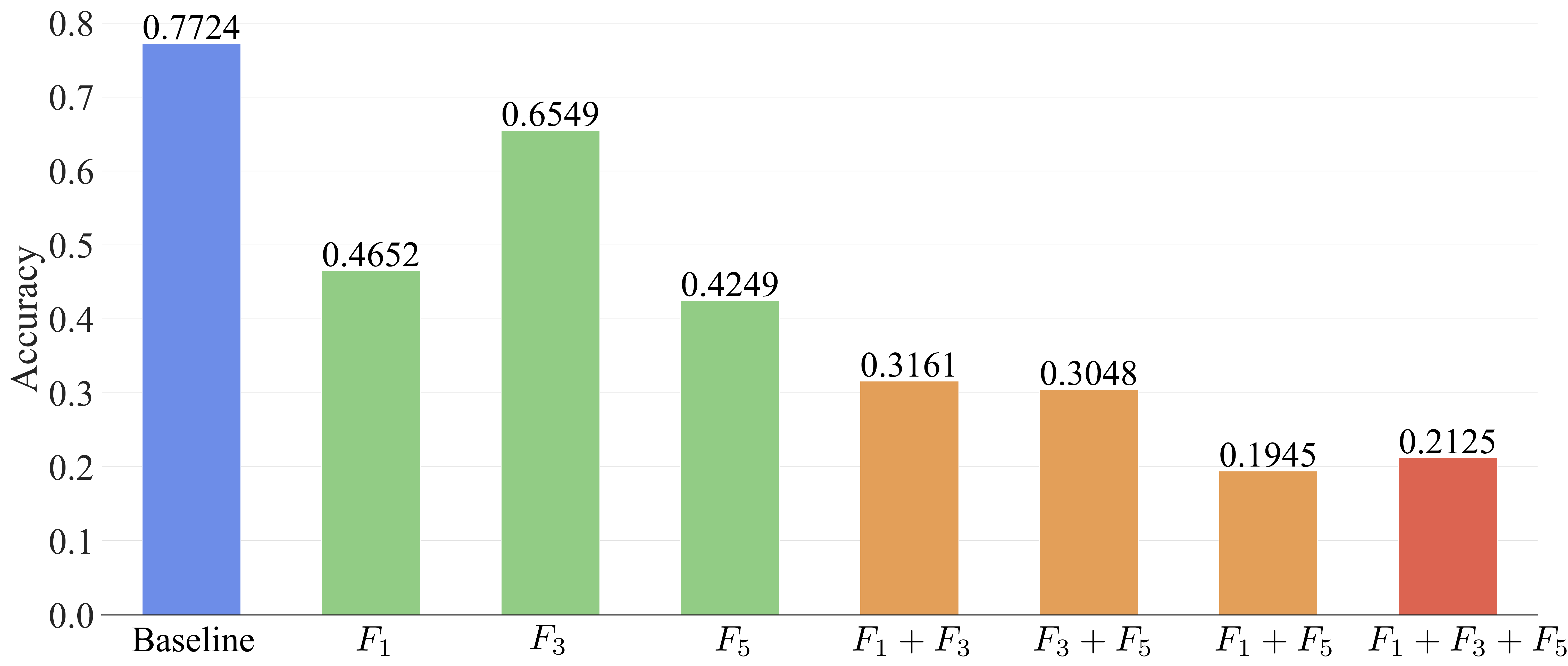}
    \caption{Impacts of combinatorial factors on an image classification DNN model}
    \label{fig:motivation2}
\end{figure*}

From the results in Figure~\ref{fig:motivation1}, different factors can cause substantial performance variances. 
To further inspect the impacts of combinatorial factors, we exercised and compared the eight executions of arbitrary combinations with three factors from Table~\ref{tab:factors} (i.e., adversarial attack ($F_1$), label flipping attack ($F_3$), and weight modification ($F_5$)) to an image recognition model trained using the CIFAR-10 dataset. 
In Figure~\ref{fig:motivation2}, the baseline model achieves 77.24\% accuracy without perturbations and we configure the settings regarding ``$F_1$'' by FGSM~\cite{ICLR2015FGSM} adversarial attack with $\sigma=0.003$, ``$F_3$'' by multi-class label flipping~\cite{NeurIPS2018flipping} on 20\% training data, and ``$F_5$'' by a random weight parameter matrix modification~\cite{Weight}. 
We could obtain two essential observations: (1) Combinatorial factor degrades the performance of DNN models, but the impacts can not be inferred by the results on individual factors (e.g., ``$F_1$'' decreases the accuracy by 30.72\% than baseline, ``$F_5$'' yields an accuracy drop of 34.75\%, ``$F_1+F_5$'' is not 65.47\% but 57.79\% instead); (2) Combinatorial factors may have some neutralized settings (e.g.,  ``$F_1+F_5$'' are slightly more effective than ``$F_1+F_3+F_5$'' to degrade the DNN accuracy).

\section{Robustness Regarding Data and Configurations}
DNN model can be represented as a function $f_{\theta}:\mathbb{X}\rightarrow\mathbb{Y}$, which accepts an input $x\in \mathbb{X} \subseteq \mathbb{R}^n$, and returns an output $y\in \mathbb{Y} \subseteq \mathbb{R}^m$, where $\theta\in\Theta$ denotes the model parameter/architecture configuration, and $\mathbb{X}$ and $\mathbb{Y}$ are the inputs and outputs in the real number domain with $n$ and $m$ dimensions, respectively. 
Formally, we can train a DNN model by: 
\begin{equation}\label{eq:trainModel}\small
    f_{\theta} = train(\mathbb{X}, \mathbb{Y}, \theta).
\end{equation}
From the above investigation, many factors exert single or combinatorial effects on the robustness performance of deep learning models in practical scenarios, e.g., recycling waste classification and traffic forecasting. According to the wide investigation in our previous work, we conclude the common robustness factors in Table~\ref{tab:factors} on two deep learning surfaces: data and configurations.
A robust DNN is expected to be resilient to a small perturbation on data $(x,y)$ and configuration $\theta$, which means the model (1) always yields the same output given the small perturbed input; (2) produces correct results in the presence of noisy labels, and (3) is stable in accuracy given perturbed model configurations. 
The following Equations~\eqref{eq1}-\eqref{eq3} define these three aspects for the DNN robustness w.r.t. data and configurations.

\begin{itemize}

\item{\textbf{Data}:} For supervised learning, a DNN model predicts the label of a sample through capturing the pattern between training data and their pre-defined labels. A trained DNN model $f_{\theta}$ is treated as robust if the input-output relation ($x,y$) always holds under perturbed inputs and outputs.
% input perturbation, irrelevant inputs and noisy labels.

\begin{itemize}
    \item \textbf{Perturbed inputs}: During the training period, perturbed inputs are commonly imposed and can mislead the learning process. In response to the perturbations, a robust DNN can be formalized as:
    \begin{equation}\label{eq1}\small
    \begin{aligned}
     & \forall x\in \mathbb{X}, \hat{x}\in \mathbb{X},\\
    & \|x-\hat{x}\|_\mathtt{p} < \sigma \Rightarrow {f_{\theta}(\hat{x})}=y\in \mathbb{Y},    
    \end{aligned}
    \end{equation}
    where $\hat{x}$ denotes the perturbed inputs under $\mathtt{p}$ normalization with $\sigma$ distance (the degree of perturbations) to the original input $x$, since the trained robust model can accept a perturbed input and return the consistent output. Adversarial attack  ($F_1$) is regarded as the different forms of $\hat{x}$.
    % TODO: 确定Table 2中的各个具体factor是否还需要写在这个部分。
    % $\hat{y}$ denotes the corresponding predicted output. 

    \item \textbf{Perturbed outputs}: If the training dataset contains corrupted or noisy output ($\hat{y}\in\hat{\mathbb{Y}}$) (e.g., under a $\delta$ distance to the ground-truth output $y$), a robust model $f_{\theta}$ can still maintain the correct prediction:
    \begin{equation} \label{eq2} \small
    \begin{aligned}
      & \forall (x, y)\in( \mathbb{X}, \mathbb{Y}), 
    \hat{y} =  y \times \tau,  \\ 
    & \|f_{\theta}(x)-\hat{y}\|_\mathtt{p} < \delta  \Rightarrow f_{\theta}(x) = y , \\
    & s.t.\ f_{\theta} = train(\mathbb{X}, \mathbb{\hat{Y}}, \theta), 
    \end{aligned}
    \end{equation}
    where $\tau\in\mathbb{R}^{m \times m}$ denotes the label transition probability matrix. 
\end{itemize}

\item{\textbf{Configurations}:} In addition to data factors, the training variance from configurations (e.g., model initialization and hyperparameters) is another reason for the robustness issues of DNNs. A trained DNN is robust if prediction variance is less susceptible to configuration perturbations:
    \begin{equation} \label{eq3} 
    \small
    \begin{aligned}
    & \forall  x\in \mathbb{X}, \theta \in \Theta, \hat{\theta} \in \Theta, \\
    & \|\theta-\hat{\theta}\|_\mathtt{p} < \eta \Rightarrow f_{\theta}(x)=f_{\hat{\theta}}(x)=y\in \mathbb{Y},       
    \end{aligned}
    \end{equation}
where $\hat{\theta}$ denotes the configuration under $\mathtt{p}$ normalization with $\eta$ distance (configuration differences) to the configuration $\theta$.

\end{itemize}

Following these definitions, we can measure DNN robustness by \emph{Cumulative confidence decision boundary} (C-CDD) through estimating the confidence of a model $f_{\theta}$ to predict all samples from $\mathbb{X}$. 
Equation~\eqref{eq:cumlativeCDD} presents a relative distance from the decision boundary to the human desired classes. 
\begin{equation}\label{eq:cumlativeCDD}\small
  \text{C-CDD} (\mathbb{X}, f_{\theta})=  \mathbb{E}_{x\in\mathbb{X}}(f_{\theta}(x)[i]-f_{\theta}(x)[j]),
\end{equation}
where $\mathbb{E}$ denotes the expectation operation, $i$ denotes the human desired class, and $j$ is the predicted class with the maximum prediction probability. 
A lower C-CDD score indicates that the model potentially has higher uncertainties in its predictions. 
\section{Search-Based Variance Prediction}
Given the two perturbation surface (i.e., data ($F_D$) and configuration ($F_C$) in Table~\ref{tab:factors}), 
we can obtain a perturbation set $T_{F_i}$ with different pre-defined values for factor $F_i \in $ \{ $F_1$, $\dots$, $F_N$ \}. Perturbation strategy $\textit{ps} \in T_{F_1}\dots\times T_{F_i}\dots \times T_{F_N}$ is a perturbation instance from all modifications given all factor combinations. Hence modified inputs, outputs, and configurations from the strategy $\textit{ps}$ can be obtained by:
\begin{equation}\label{eq:perturb_train}
    \small
    \begin{aligned}
  \hat{\mathbb{X}},\hat{\mathbb{Y}}, \hat{\theta} = perturb_{\textit{ps}}(\mathbb{X},\mathbb{Y},\theta).
    \end{aligned}
\end{equation}

\begin{figure}
    \centering
    \includegraphics[scale = 0.95]{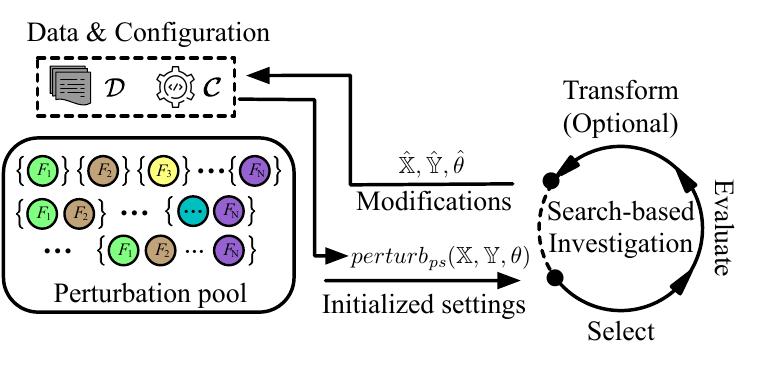}
    \caption{Search-Based Investigation Framework}
    \label{fig:task1frame}
\end{figure}

\begin{table}[!t]
    \centering
    \caption{Robustness affecting factors}
    \begin{tabular}{cll}
    \toprule
        \textbf{Surface} & \textbf{Factor} & \textbf{Target}\\\hline
        \multirow{4}{*}[-0.5\dimexpr \aboverulesep + \belowrulesep + \cmidrulewidth]{\makecell{ Data \\ ($F_D$) }} & $F_1$ Adversarial attack &  \multirow{2}{*}{Input $\mathbb{X}$} \\
                        & $F_2$ Out of distribution  &        \\\cline{2-3}
        
                        &$F_3$ Label flipping attack & \multirow{2}{*}[-0.5\dimexpr \aboverulesep + \belowrulesep + \cmidrulewidth]{Output $\mathbb{Y}$} \\ 
                        &$F_4$ Label noise injection  \\ \hline
        \multirow{6}{*}[-0.5\dimexpr \aboverulesep + \belowrulesep + \cmidrulewidth]{\makecell{ Configuration \\ ($F_C$)}}& $F_5$ Weight modification   & \multirow{6}{*}{Model $f_{\theta}$ } \\ 
                    & $F_6$ Bias modification &  \\
                       & $F_7$ Conv layer modification&  \\
                       & $F_8$ FC layer modification & \\
                       & $F_9$ Number of threadings & \\
                       & $F_{10}$ Pseudorandom seeds & \\
                       &$\dots$ &$\dots$ \\
   \bottomrule
    \end{tabular}
    \label{tab:factors}
\end{table}
Given $N$ factors with each having $T$ modifications to the data and (or) configurations, we will have combinations of $2^{T*N}$ perturbation strategies to evaluate the robustness of a DNN model.
% This combinatorial optimization problem is NP-hard to find the optimal target from a finite set.
 Inspired by search-based software engineering~\cite{chen2018sampling}, we present a search-based framework to conduct the \emph{selective optimization} and reduce the searching space as in Figure~\ref{fig:task1frame}. 
The perturbation pool collects all perturbation strategies, and we will select several strategies as the initialization settings for the investigation module. Here, the selected strategies are utilized to generate perturbed data samples and configuration settings via Equation~\eqref{eq:perturb_train}. 
The investigation module contains three main procedures, i.e., select, evaluate and transform (optional), which can be leveraged for multiple solutions to the combinatorial optimization problem. 
(1) Select. The framework selects the candidate strategies from the perturbation pool. 
(2) Evaluate. The performance variances from the selected strategies are evaluated in this step. 
(3) Transform (optional). After the evaluation, some strategies with satisfactory results may be further transformed to jump out of the local search space. Here, the transformed strategies and the original satisfactory strategies are mixed for the next round of selective optimization.  
For this search-based investigation framework, the optimal $\textit{ps}^{\star}$ refers to finding the largest performance variances $PV(\textit{ps})$:
\begin{equation}\label{eq:opt_objective}\small
\begin{aligned}
    \textit{ps}^{\star} &= \max_{\textit{ps}}PV(\textit{ps}) \\
    &= \max_{\textit{ps}}|\text{C-CDD}(\hat{\mathbb{X}},f_{\hat{\theta}})-\text{C-CDD}(\mathbb{X},f_{\theta})|,
    \end{aligned}
\end{equation}
where $\hat{\mathbb{X}}$ and $\hat{\theta}$ are derived from Equation~\eqref{eq:perturb_train}.

%The investigation module balances the cost and satisfaction of the user's specifications and engineering requirements. 
Our proposed framework is highly extensible and can accommodate the mainstream search-based approaches by fitting the objective function in Equation~\eqref{eq:opt_objective}. The representative mainstream approaches, including evolutionary algorithm (EA)~\cite{awad2021dehb}, reinforcement learning (RL)~\cite{mazyavkina2021reinforcement}, sequential model-based optimization (SMBO)~\cite{candelieri2019sequential}, geometric learning (GL)~\cite{chen2018sampling} can share the same steps but may occur in different orders. 

\textbf{Evolutionary algorithm (EA)}
EA is a fast global optimization algorithm, inspired by biological terminology, whereas an initial set of individual solutions are introduced to small random changes and updated iteratively. 
Here, we use EA to search from the perturbation pool for the expected $\textit{ps}^{\star}$. 
EA performs the iterative way in the investigation module, and for $\forall \textit{ps}_i^{g+1}$ in $(g+1)$-th generation, the transform step is divided as mutate and crossover:
\begin{enumerate}
    \item [(1)] Select. Firstly, three strategies $\textit{ps}_{r_1}^g$, $\textit{ps}_{r_2}^g$, $\textit{ps}_{r_3}^g$ are randomly selected from the candidate strategies in $g$-th generation, where the subscript of the strategy ($r$) indicates the index of the candidate in the current generation. The indices of these three strategies should satisfy: $r_1 \neq r_2 \neq r_3 \neq i$, where $i$ is the index of the expected strategy $\textit{ps}_i^{g+1}$.
    \item [(2)] Mutate. A candidate $\textit{ps}_i^{g+1}$ can be obtained by the mutation operation: $\textit{ps}_i^{g+1} = \textit{ps}_{r_1}^g + \epsilon (\textit{ps}_{r_2}^g-s_{r_3}^g)$, where $\epsilon\in(0,1]$ denotes a scaling coefficient for the mutation proportion.
    \item [(3)] Crossover. The strategy $\textit{ps}$ can be treated as a $K$-length vector that contains a sequence of modification methods. Let $\textit{ps}_i^g[j]$ denotes the $j$-th ($1\leq j \leq K$) modification in the perturbation strategy. The crossover follows a random probability value from the uniform distribution, i.e., $r_i^{g}[j] \sim U(0,1)$. The $(g+1)$-th generation of each particular modification can be obtained as: 
    \begin{equation}\small
    \textit{ps}_i^{g+1}[j] = \left\{ 
     \begin{array}{ll}
           \textit{ps}_i^{g}[j], & \mathrm{if}\ r_i^{g+1}[j] < p_r \\ 
           \textit{ps}_i^{g+1}[j],& \mathrm{otherwise},
     \end{array}
     \right. 
    \end{equation}
    where $p_r$ is a user-specified replacement rate (e.g.,  $0.9$).
    \item [(4)] Evaluate. For each newly generated strategy $\textit{ps}_i^{g+1}$, we produce a different DNN model by Equation~\eqref{eq:trainModel} and return the strategy with a lower C-CDD score from $\{\textit{ps}_i^{g}, \textit{ps}_i^{g+1} \}$. 
\end{enumerate}

EA can balance the cost and satisfaction from the user's specifications and engineering requirements, by configuring the value of $\epsilon$ and $p_r$ to control how drastically it reaches a fitness degree constrained by Equation~\eqref{eq:opt_objective}.

\textbf{Reinforcement learning (RL)} 
The combinatorial optimization problem can be modeled as a sequential decision-making process (DMP). Among the solutions to DMP, RL is a reward-based one for finding the desired strategy. The agent from RL conducts like humans to observe a series of positive or negative rewards during the state transition process $T$. The observation is from the state $S$, representing the robustness of the model with parameter $\hat{\theta}$ to transmit to the next state according to a learned action $A$. Then the agent gradually obtains more observed experiences and performs better to reach the objective. Here, the selection of the action for a specific state is guided by a reward $R$ from the robustness variance. 
The module in this context is transferred as a Markov Decision Process $MDP = \langle S, A, R, T, \gamma\rangle$. We elaborate on each of the tuple elements: 
%detail the action procedures 
\begin{itemize}
    \item [(1)] State ($S$). We represent the network state $s\in S$ through a pair of features, $s=\langle f_{\hat{\theta}}, \mathrm{C-CDD}\rangle$ for the newly produced DNN model based on Equation~\eqref{eq:perturb_train} and corresponding cumulative confidence score. 
    \item [(2)] Action ($A$). During the learning process, the configurable settings from the strategy are selected from the perturbation pool, whereas $a\in A$ denotes the changing settings. 
    \item [(3)] Transition ($T$). $T_a(s,s')=T_{a}(s_{t+1}=s'\ |\ s_t = s) $ denotes the probability of the transition from $s$ to $s'$ under action $a$ at $t$ time (the $t-$th selection of the strategy).
    \item [(4)] Reward ($R$). RL acts in MDP is to find a policy $ A \times S \rightarrow [0,1] $ that maximizes the expected cumulative reward, and $r\in R$ denotes the immediate rewards from $t$ to $t+1$ transition time.
    \item [(5)] Scaling rate ($\gamma$). A lower $\gamma$ value contributes more weight to the short-term reward, whilst the higher $\gamma$ value reflects the learning actions would reward more towards long-term ones. 
\end{itemize}
With the aforementioned instances, a set of strategies are firstly selected and evaluated by C-CDD. The agent receives and learns the numerical rewards from the transitions, which can be formed as a sequence, i.e., $s_0, a_0, r_1, s_1, a_1, r_2, \cdots$. Then the sequence can be utilized in the learning iteration until convergence and reaching our defined objective.

\textbf{Sequential model-based optimization (SMBO)} 
SMBO is another alternative option in addressing the problems regarding searching for configurable perturbation strategies. 
Our framework is interchangeable to fit with the components and facets, that approximate the optimal solution based on a sequential model. 
SMBO utilizes prior belief distribution to fit a Gaussian mixture model for the desired strategy to DNN model $f_{\theta}$. 
Specifically, for the search-based variance prediction, SMBO conducts two steps for the estimation of the global optimal ${ps}^{\star}$ with the largest performance variance:
\begin{enumerate}
    \item [(1)] Evaluate. SMBO incrementally accepts from $0$ to $t$ number of distinct perturbation strategies and constructs an acquisition function using the posterior for $f_{\theta}$ with the previous $t-1$ evaluations $\{({ps}_i,PV({ps}_i)\}_{i=1}^{t-1}$. 
    \item [(2)] Select. The acquisition function selects the next potential perturbation strategy ${ps}_{i+1}$ and calculates the corresponding $PV({ps}_{i+1})$ for the update of the acquisition function. The acquisition function determines the next points (another strategy) worth being evaluated, which keeps the balance between different perturbation factors (in the sense of which factor is more essential) and exploration of the strategies (in the sense of which settings are more effective to degrade). 
\end{enumerate}

The fitness agreement is still determined by humans, when SMBO reaches a given value or a specific threshold time (i.e., total number of evaluation times $t$), the potential strategy from the current acquisition function is regarded as the approximation of the optimal settings. 

\textbf{Geometric learning (GL)}
GL decides the next sampling point by considering the shape of the sample. For a given huge set of candidate strategies (e.g., 1000 strategies) in the perturbation pool, the typical method SWAY finds the optimal solution based on the divide-and-conquer method in three steps:
\begin{enumerate}
    \item [(1)] Divide. Divide the candidate strategies into two subsets, ``west side'' and ``east side'' if the remaining strategy size is bigger than a given threshold else return the remaining candidates.
    \item [(2)] Conquer (Evaluate). Train the model with the representatives ${ps}_{W}$ and ${ps}_{E}$ in each set and output the corresponding $PV({ps}_{W})$ and $PV({ps}_{E})$.
    \item [(3)] Prune. Compare $PV({ps}_{W})$ and $PV({ps}_{E})$, prune the candidate set with lower performance, and jump to step (1) for the optimization of the other candidate subset.
\end{enumerate}
In GL, both the ``Divide'' and ``Prune'' steps are for the selection in the search-based investigation framework.

\section{CONCLUSION}
In this article, we present and study the two main sources of performance variances in deep learning, i.e., imperfect data and configurations. 
We have showcased the performance variances in real-world applications in a range of CSIRO's missions. 
The proposed framework sheds light on future research in robust deep learning by providing a promising ground for predicting influential factors with search-based optimizations. Furthermore, it is promising to explore the potential solutions for mitigating the combinatorial effects from multiply robustness-affecting issues. In future work, we would like to incorporate iterative optimization into the related countermeasures, where we could optimize each subproblem iteratively.

%\section{ACKNOWLEDGMENT}

%biographical info:
% [Full name] is [role] at [institution] at [city, state, postal code, country]. [His/Her] research interests include [3 very brief (not a complete list of) topics]. [Last name] received [his/her] [highest degree] in [topic] from [institution]. [He/She] is a [member/fellow/other] at [professional organization]. Contact [him/her] at [website or email address].

\begin{IEEEbiography}{Guanqin ZHANG} is a Ph.D. student at the University of Technology Sydney, Australia. His research interests include artificial intelligence, software engineering, and program verification. Contact him at Guanqin.zhang@student.uts.edu.au.
\end{IEEEbiography}

\begin{IEEEbiography}{Jiankun Sun} is working towards the PhD degree at the University of Science and Technology Beijing, Beijing, China. He is currently a visiting student with the School of Computer Science, Faculty of Engineering and Information Technology, University of Technology Sydney. His research interests include software engineering, artificial intelligence, and security. Contact him at sunjk@xs.ustb.edu.cn.
\end{IEEEbiography}

\begin{IEEEbiography}{Feng Xu} is a Master of Science (Research) major in Computer Science student at the University of Technology Sydney, NSW Australia. His research interests include software engineering, program analysis, and artificial intelligence. Contact him at feng.xu@student.uts.edu.au.
\end{IEEEbiography}

\begin{IEEEbiography}{H. M. N. Dilum Bandara} received the M.S. and Ph.D. degrees in Electrical and Computer Engineering from Colorado State University, USA. He is a Senior Research Scientist with the Architecture and Analytics Platforms Team, Data61, CSIRO. He is also a Conjoint Senior Lecturer at the School of Computer Science and Engineering, University of New South Wales (UNSW). His research interests include Distributed Systems (Blockchain, Cloud, and P2P), Computer Security, Software Architecture, IoT, and Data Engineering, as well as multidisciplinary applications of those technologies. He is a Senior Member at IEEE and a Charted Engineer at IESL.
\end{IEEEbiography}

\begin{IEEEbiography}{Yulei Sui} is currently an associate professor with the
School of Computer Science, University of Technology Sydney. His research interests broadly include program analysis, software engineering, and security. He was a plenary talk speaker at EuroLLVM 2016. He is the recipient of a 2020 OOPSLA Distinguished Paper Award, a 2019 SAS Best Paper award, a 2018 ICSE Distinguished Paper Award, a 2013 CGO Best Paper Award, an ACM CAPS Award, and an ARC Discovery Early Career Researcher Award (2017–2019)
\end{IEEEbiography}

\begin{IEEEbiography}{Shiping Chen} (Senior Member, IEEE) is a Senior Principal Research Scientist with CSIRO Data61. He also holds a conjoint Professor with UNSW, UTS and Macquarie University. He has been working on distributed systems for over 20 years with a focus on performance and security. He has 280+ publications in these areas. He is active in the research community through journal editorships and conference PC/Chair services. His current research interests include application security, blockchain, and digital services. He is a Fellow of the Institute of Engineering Technology (FIET)
\end{IEEEbiography}

\begin{IEEEbiography}{Tim Menzies} (Fellow, IEEE) received the PhD, degree from UNSW Sydney in 1995. He is currently a professor in computer science with NC
State University, USA, where he teaches software engineering, automated software engineering,
and programming languages. His research interests include software engineering (SE), data mining, artificial intelligence, search-based SE, and open-access science
\end{IEEEbiography}

\end{document}